\documentclass{article}
\usepackage{spconf,amsmath,graphicx}
\usepackage{makecell}
\usepackage{xcolor}
\usepackage[table]{xcolor}
\usepackage[T1]{fontenc}
\usepackage{caption}
\usepackage{subcaption}
\usepackage{placeins}

\title{PATCH ENSEMBLES FOR ROBUST SALMON RE-IDENTIFICATION WITH WEAK TRAJECTORY LABELS}
\name{
Espen Uri Høgstedt$^1$,
Christian Schellewald$^2$,
Annette Stahl$^3$,
Rudolf Mester$^1$
\thanks{This work was supported by the Research Council of Norway (RCN) under project number 344022, titled \textit{Computer Vision and Artificial Intelligence-based Salmon Identification and automated long-term welfare assessment in aquaculture net-pens (cAIge)}.}
\thanks{\copyright 2026 IEEE. Published in the 2026 IEEE International Conference on Image Processing (ICIP). Personal use of this material is permitted. Permission from IEEE must be obtained for all other uses, in any current or future media, including reprinting/republishing this material for advertising or promotional purposes, creating new collective works, for resale or redistribution to servers or lists, or reuse of any copyrighted component of this work in other works.}
}

\address{
    $^{1}$Department of Computer Science, 
    Norwegian University of Science and Technology, Norway\\
    $^{2}$SINTEF Ocean, Norway\\
    $^{3}$Department of Engineering Cybernetics, 
    Norwegian University of Science and Technology, Norway
}
\begin{document}
%
\maketitle
\begin{abstract}
Salmon re-identification in commercial net-pens is challenging due to large populations, which impose strict accuracy requirements and make large-scale labeled data acquisition infeasible.
Trajectory IDs can be used as proxy labels, but this introduces trajectory-ID bias.
To address these challenges, we propose a patch-based re-identification framework that fuses patch-level predictions into a salmon identity decision.
A key component is the prediction of the salmon's lateral line, enabling extraction of texture-anchored patches and patch slices.
To enable realistic evaluation, we introduce an experimental setup using multiple cameras placed 6 m apart, allowing the same fish to be recorded in different trajectories. This enables the construction of a cross-camera test set through manual match confirmation.
Our ensemble approach outperforms the full-image baseline in same-trajectory validation (0.932 to 0.965 mAP) and cross-camera testing (0.609 to 0.860 mAP). The substantial improvements in the cross-camera setting demonstrate improved generalizability and robustness. Code and data: https://github.com/espenbh/salmon-reid-patch-ensemble.

\end{abstract}
\section{Introduction}
\label{sec:intro}

Monitoring individual salmon over long periods will enable researchers to study how individuals develop over time, and aquaculture operators to tailor operations to the needs of each salmon. Today, fish re-identification relies on electronically readable tags attached to fish, which is effective for a small number of fish, but is costly, invasive, and infeasible for the hundreds of thousands of fish present in commercial net-pens.

Recent progress in deep learning-based computer vision is opening new opportunities for non-invasive salmon re-identification directly from images. Such methods would scale naturally, require no human intervention once deployed, and avoid stressing the fish. Earlier approaches have demonstrated promising performance on small numbers of individuals or under controlled laboratory conditions \cite{Hogstedt25, cisar2021}. However, successful salmon re-identification in large commercial net-pens remains unsolved.

\begin{figure}
    \centering
    \includegraphics[width=\linewidth]{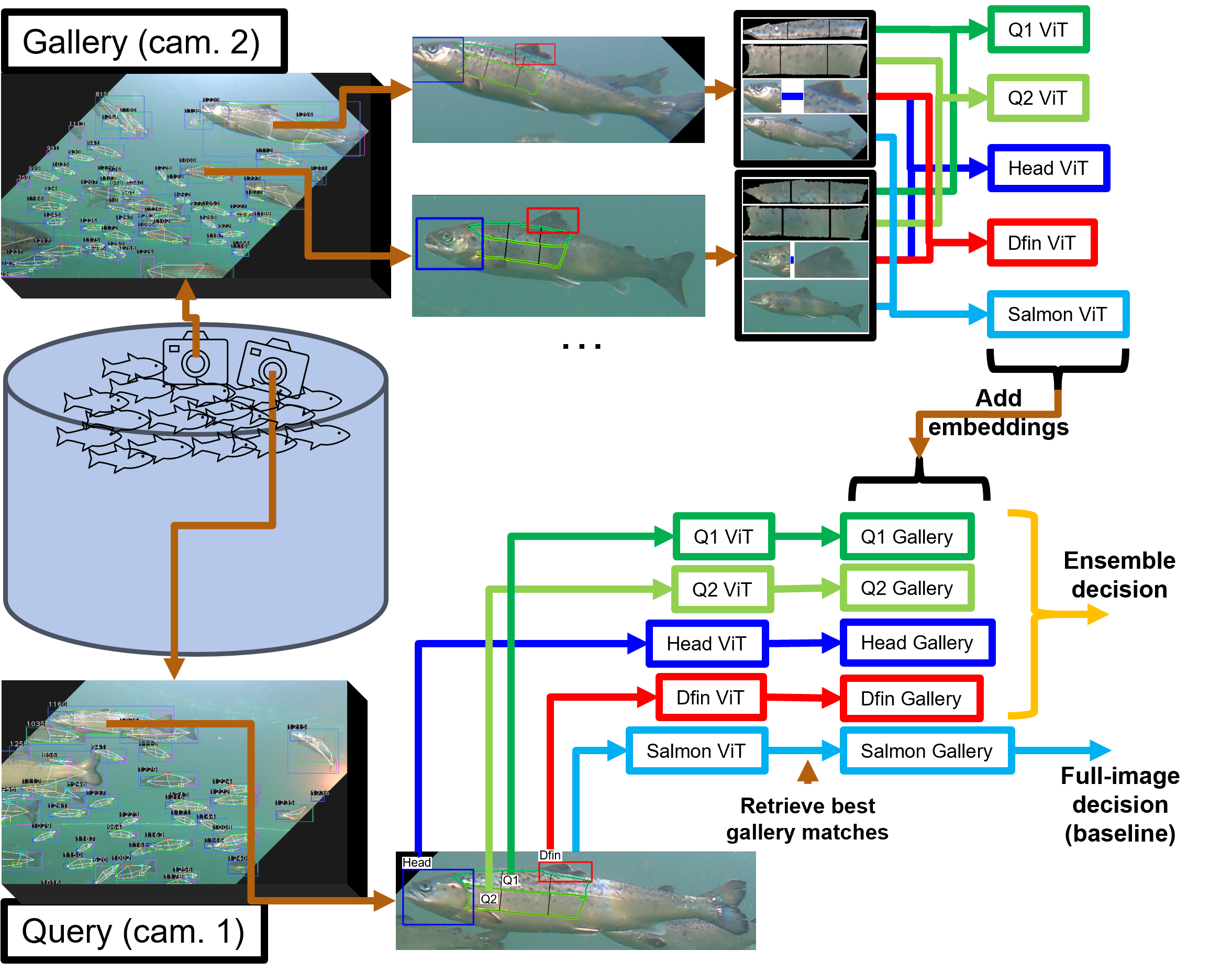}
    \caption{Overview of the salmon re-identification test pipeline. Images are extracted from videos recorded by two different cameras. Each salmon ROI (region of interest) is divided into four patches (head, dorsal fin, Q1, Q2), with Q1 and Q2 further divided into three slices using the predicted lateral line. Each patch is processed by a dedicated embedding network, and patch-level matches are fused into a full-fish decision.}
    \label{fig:complete_pipeline}
\end{figure}

Large-scale salmon re-identification has two central challenges. First, the accuracy requirements are very high, with a single pen containing up to 400,000 visually distinct salmon sides (200,000 individuals with two sides each). Second, acquiring labeled data at this scale is infeasible. Humans cannot manually match individuals across hundreds of thousands of identities, and tagging all fish is prohibitively expensive.

One way to address the lack of labeled data is to use trajectory IDs as proxy labels for salmon identity. This, however, introduces a new and challenging problem: trajectory-ID bias. Since each fish is associated to a single trajectory with similar lighting conditions, models trained on such labels struggle to recognize the same individual across different trajectories. 

To mitigate this bias and reduce the accuracy demands of our embedding networks, we propose a re-identification framework where each salmon ROI (region containing a single salmon) is split into multiple patches, and a dedicated re-identification pipeline is applied per patch. The patch-level predictions are then fused into a single salmon-level decision. 
This approach relaxes the accuracy requirements of individual embedding networks, since an error in one patch can be compensated for by correct predictions in the others. 
In addition, a single patch is easier to augment realistically than the full salmon ROI, as the patch is more planar, has more uniform lighting, and exhibits less deformation.

While this ensemble approach addresses algorithmic development, one challenge remains: how to evaluate re-identification performance. To address this, we design an experimental setup using multiple cameras placed relatively closely together within the net-pen, so that the same fish may be recorded in different trajectories within a short time span. This configuration allows manual verification of matches proposed by the algorithm. As a result, we obtain realistic evaluation data that includes cross-trajectory variation while keeping the manual annotation effort manageable.

Figure \ref{fig:complete_pipeline} illustrates our proposed re-identification pipeline. We begin by tracking salmon in two videos recorded from different cameras using the method in \cite{Hogstedt2025tracking}, and extracting a subset of high-quality salmon ROIs from the generated tracks. Each salmon ROI is then divided into four patches: head, dorsal fin, Q1 (anteriodorsal body quarter), and Q2 (anterioventral body quarter). The head and dorsal fin are obtained directly from the tracker, while Q1 and Q2 are obtained by a segmentation model. From the segmented regions (Q1, Q2), we extract the predicted lateral line and use it to further divide each body quarter into three texture-anchored slices. Each patch (patch + three slices for the body quarters) is processed by a separate embedding network. 
Next, we build a separate gallery for each patch type using images from camera 2. For each salmon ROI in camera 1 (query image), we compare its patch embeddings to the corresponding galleries, and aggregate the resulting patch-level scores into one similarity score per gallery sample. Finally, we evaluate performance by manually verifying the retrieved matches and computing mean average precision (mAP) from these verified ground-truth pairs.

Our results show that the cross-camera test setup is substantially more challenging and realistic than the single-trajectory validation set, with the full-image baseline (model 1, Table \ref{tab:res}) dropping 0.323 mAP from validation to test. Moreover, our ensemble method outperforms the full-image baseline on both the validation (0.932 to 0.965 mAP) and test (0.609 to 0.860 mAP) sets, showing especially strong improvements on the test set, demonstrating improved robustness and generalizability.

Our contributions are: (i) A method for predicting the salmon lateral line and using it to extract texture-anchored body patches and slices. (ii) An ensemble approach that improves the robustness of models trained on trajectory proxy labels by mitigating trajectory-ID bias. (iii) An experimental setup for realistic evaluation of re-identification methods in commercial net-pens.

\section{Related work}
\label{sec:related_work}

Fish re-identification studies typically build on well-established methods adapted from larger domains, such as vehicle and person re-identification. Existing approaches include  classification models with one class per identity \cite{al2018, Hogstedt25}, metric-learning methods that train networks to generate discriminative embeddings 
\cite{zhou2022, mathisen2020, shi2024},
and classical techniques such as keypoint matching \cite{pedersen2022, dala2016}. Another line of work leverages the dot pattern on salmonids, either by comparing dots or manually constructed dot embeddings directly \cite{debicki2021, cisar2021, stien2017}, or by providing dot segmentation masks to a deep neural network \cite{zhou2022}. To our knowledge, all learning-based methods, with one exception, rely on ground-truth identity labels obtained through tagging 
\cite{debicki2021, cisar2021, stien2017, dala2016, shi2024},
manual or semi-automated labeling 
\cite{pedersen2022, Hogstedt25},
or experimental setup with isolated fish \cite{cisar2021, al2018}. The exception is the work by Mathisen et al. \cite{mathisen2020}, which uses trajectory IDs as proxy labels. These labels were taken as ground-truth, and did not address trajectory-ID bias or cross-trajectory generalization. In contrast, our method is designed to address the limitations of weak trajectory labels by combining an ensemble architecture with a multi-camera evaluation setup.

In addition to differences in learning paradigms, existing works also vary in the image regions they use for identification. Several studies isolate specific regions (such as the head \cite{mathisen2020}, the anterior half \cite{Hogstedt25}, a specified bounding box \cite{cisar2021} or the parr mark region \cite{shi2024}). However, no prior work in fish re-identification extracts multiple informative patches and combines them in an ensemble, nor does any method segment texture-anchored patches by leveraging the salmon lateral line.

Together, these gaps motivate our approach: a lateral line-anchored patch extraction method, an ensemble of complementary patch-level identity predictions, and a multi-camera evaluation setup that enables realistic assessment of re-identification performance in commercial net-pens.

\section{Method}
\label{sec:method}
In this section, we describe our approach in the order of the data processing pipeline. 
We first outline the raw data collection procedure and the construction of the re-identification dataset (\ref{sec:method_data_collection}–\ref{sec:method_reid_dataset}), then describe the re-identification pipeline (\ref{sec:method_reid_data}-\ref{sec:method_architectures}), and finally present the ensemble method (\ref{sec:method_fusion}) and the evaluation protocol (\ref{sec:method_eval}).

\subsection{Data Collection, Fish Tracking and Initial Filtering}\label{sec:method_data_collection}
Video data were collected at Korsneset on 18 April 2024 using six GoPro cameras, each attached to a line and lowered along the interior net wall of a commercial salmon pen. Cameras were spaced 6\,m apart and positioned at a depth of 2 m (Fig. \ref{fig:complete_pipeline} illustrates two of the cameras). Each camera recorded approximately 30 minutes of video at 30 fps, stored in clips of 15,930 frames at a resolution of $2160\times3840$ pixels. The same procedure was repeated in two net-pens.

Subsequently, salmon and body-part bounding box tracks were extracted from the raw videos using the multi-object tracking approach from \cite{Hogstedt2025tracking}. We removed salmon detections with diagonal length below $l_{\mathrm{diag}}$, those with occluded body parts, and all tracks shorter than $min\_traj\_length$ consecutive frames. For each fish, only its longest uninterrupted track was retained.

\subsection{Body Quarter Segmentation}
\label{sec:method_seg_dataset}
To create a dataset for salmon body quarter segmentation, we manually annotated salmon images from a representative video clip from camera 1 in net-pen 2 (frames 10,800 to 11,400). This clip had an appropriate fish density, with individuals positioned at suitable distances from the camera. Images were extracted using $l_{diag} = 250$ and $min\_traj\_length = 10$ (Section \ref{sec:method_data_collection}), and sampled at 5 to 10 frame intervals to balance annotation effort, number of labeled identities and per-fish viewpoint variation.

\begin{figure}
    \centering
    \includegraphics[width=\linewidth]{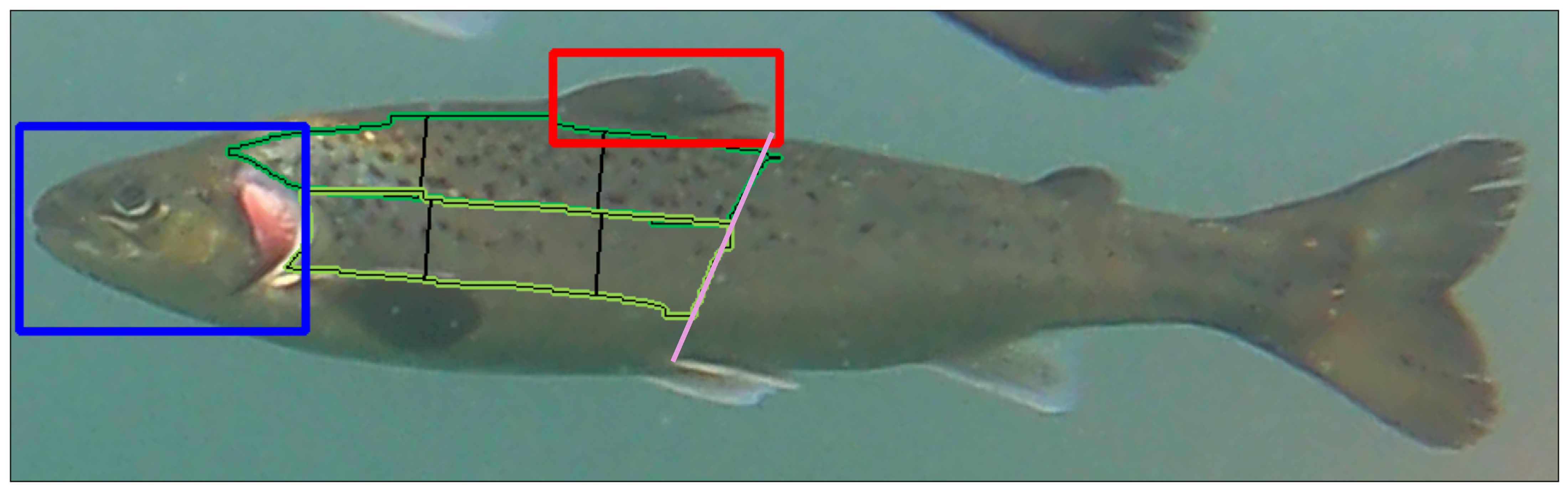}
    \caption{Aaron with predicted segmentation masks and tracker body parts. Q1 in green, Q2 in olive, dorsal fin in red, head in blue, and black lines separating the patch slices. Additionally, the backline is drawn in the image as a pink line.}
    \label{fig:Aaron}
\end{figure}

Only fish with a visible lateral line in at least one frame were considered. Once identified in a single frame, the lateral line was inferred in the remaining frames using dot-pattern cues. This lateral line served as the separating boundary between Q1 and Q2. The following structures were annotated (Figure \ref{fig:Aaron}):  
(i) the backline (dorsal fin posterior margin to pelvic fin root),  
(ii) Q1 (anteriodorsal body quarter),  
(iii) Q2 (anterioventral body quarter), and 
(iv) the operculum.  
One trajectory ID (5 images) was used for validation, while the remaining 71 IDs (401 images) formed the training set, ensuring a high number of annotated samples available for learning.

A yolov8l Ultralytics model \cite{yolov8} was trained to segment Q1, Q2, and the operculum using the created dataset. Although the operculum was not used downstream, we included it as an auxiliary class during training to improve the model robustness. The backline was used only to define boundaries of Q1 and Q2. Default hyperparameters were used except rotation ($30^\circ$), perspective augmentation (0.001), and mask downsampling ratio (1). The model was trained for 50 epochs with a batch size of 8 and an image size of 640, achieving a best validation mAP$_{50-95}$ of 0.863 (Q1) and 0.796 (Q2).

\subsection{Slice Extraction}\label{sec:method_slice}
The four corners of each body quarter were found as support points of the convex hull of the segmentation mask, extracted in directions offset by $\pm 70^\circ$ (determined empirically) from the swimming direction (calculated from the head and tail fin bounding boxes). The two points spanning the lateral line of the salmon (ventral edge of Q1 and dorsal edge of Q2) were then employed to split the body quarter into three slices by first rotating the body quarter such that the lateral line was oriented horizontally, and then placing cuts at 0.3 and 0.7 of the midline length (Figure \ref{fig:Aaron}).
Afterward, the slices were expanded such that adjacent slices overlapped by two tokens. This ensured that the content located near the slice centers remained visible even with small misalignments.

\subsection{Re-Identification Dataset Construction}\label{sec:method_reid_dataset}
We assembled the re-identification dataset using the first clip from cameras 1 and 2 in net-pen 1. 
Images were extracted using $l_{diag}=600$ and $min\_traj\_length=20$ (Section \ref{sec:method_data_collection}), and every 5th frame of each trajectory was included. 
A sample was kept if segmentation succeeded for both Q1 and Q2, and if both resulting masks contained more than 25\% foreground pixels after cropping each mask to its smallest enclosing rectangle. The training set was assembled from frames 750-14,750 (processed in blocks of 1000 frames) of camera 1 (12,539 images, 923 IDs), the validation set from frames 14,750-15,930 of camera 1 (1289 images, 84 IDs), and the test set from frames 11,410-12,590 of camera 2 (1462 images, 101 IDs). The test and validation sets were temporally aligned by determining the start frame of the test set based on a reference fish (Aaron, see Figure \ref{fig:Aaron}) manually observed in both cameras.

\subsection{Data Pipeline and Augmentation}\label{sec:method_reid_data}
For each sample (salmon ROI), we first applied a perspective augmentation to the full image, followed by rotation and cropping of Q1 and Q2. The three slices from each quarter were then extracted, and full-image, patch-level and slice-level pixelwise augmentations were applied independently. Perspective augmentation was implemented using OpenCV \cite{Bradski2000}, while pixelwise augmentations were applied using the Albumentations library \cite{Buslaev2018}. Slice images were padded and resized to $224\times224$ pixels prior to augmentation. After augmentation, all images were processed using the DINO preprocessing pipeline \cite{Simeoni2025} before being fed to the models.

A curriculum schedule controlled the augmentation difficulty:  
Epoch 0: no augmentation;  
Epochs 1-4: added perspective transformations (scale 0.05), brightness/contrast perturbations, and tone curve transformations;  
Epochs 5-6: increased the scale parameter of the perspective transformation to 0.15 and added motion blur and HSV perturbations;  
Remaining epochs: additionally introduced median blur and illumination variation.

\subsection{Network Architectures and Training}\label{sec:method_architectures}
All models used projection heads composed of four layers in sequence: linear, GELU, dropout, and linear. We evaluated two model families.  
\textbf{Single-patch models:} A DINOv3-base backbone \cite{Simeoni2025} followed by a projection head with a 512-dimensional hidden layer.  
\textbf{Sliced-patch models:} Quarter and slice inputs are passed through a shared DINOv3-base backbone (via separate forward passes) and then through their individual projection heads. The resulting four embeddings are concatenated and processed by a fusion head with a 1024-dimensional hidden layer. Losses are calculated for both the individual patch/slice heads and for the fusion head.

Two full-image models (single-patch models) were trained using the complete salmon ROI as input, one with a DINOv3 ConvNeXt backbone and one with a DINOv3 vision transformer (ViT) backbone. The full-image ViT model serves as the baseline. In addition to these, we trained single-patch models for each patch type (Q1, Q2, head and dorsal fin) and sliced-patch models for Q1 and Q2, all using a DINOv3 ViT backbone. In all architectures, the first half of the backbone parameters (excluding LayerNorm) was frozen.

Models were trained using a weighted combination of SmoothAP \cite{Brown2020}, ArcFace \cite{Deng2019} (margin 0.3, scale 30), and Multi-Similarity \cite{Wang2019} losses, as implemented in the PyTorch Metric Learning library \cite{Musgrave2020}. The Multi-Similarity \cite{Wang2019} miner followed the curriculum (epsilon $0.1 \rightarrow 0.08 \rightarrow 0.07 \rightarrow  0.05$; number of sampled IDs $8 \rightarrow 16 \rightarrow 16 \rightarrow  32$). We trained for 20 epochs using AdamW \cite{Loshchilov2017} (weight decay 0.02), batch size 128, image size $224\times224$, cosine-annealing with warm restarts \cite{Loshchilov2016} (T$_0$=$\frac{1}{16}$, T$_{mult}$=$2$), gradient clipping (max-norm 5), and mixed-precision training with GradScaler. Loss weights were 1 for Multi-Similarity, 1 for SmoothAP, and 0.1 for ArcFace.

\subsection{Ensemble Method}
\label{sec:method_fusion}
We fused patch-level predictions into salmon-level decisions by computing a similarity value between each query ($i$) and gallery ($j$) salmon ROI.
The fused similarity scores $\hat{S}_{ij}$ were computed as a weighted sum of per-patch reciprocal rank scores $rr^{(p)}_{ij}$ and temperature-scaled embedding similarities $s^{(p)}_{ij}$, with $p$ indexing patch types (Equation \ref{eq:fusion}). This yields final scores that capture both robust rank-based signals and detailed similarity information. 
We set $\lambda = 0.75$, $\tau = 0.7$ and $k = 20$. The exact parameter settings are less important, as performance remains stable across a broad range of settings (sensitivity analyses provided in the supplemental materials).

The temperature-scaled similarity $s^{(p)}_{ij}$ is computed by first calculating $\tilde{s}_{ij}^{(p)} $ as shown in Equation \ref{eq:embedding_similarity}, and then min-max normalizing these scores independently for each query. $cos \: \theta_{ij}^{(p)}$ is the cosine similarity between query patch embedding $i$ and gallery patch embedding $j$. The reciprocal rank score $rr^{(p)}_{ij}$ is computed as shown in Equation \ref{eq:rr}, where $r^{(p)}_{ij}$ is the rank of gallery patch $j$ retrieved by query patch $i$.

\begin{subequations} \label{eq:fusion_block} \begin{align} 
\hat{S}_{ij} &= \lambda \sum_{p} rr^{(p)}_{ij} + (1 - \lambda) \sum_{p} s^{(p)}_{ij}, \label{eq:fusion} \\
\tilde{s}_{ij}^{(p)} &= \exp\!\left( -(1 - \cos \theta_{ij}^{(p)})/\tau \right), \label{eq:embedding_similarity} \\
rr^{(p)}_{ij} &= \frac{1}{k + r^{(p)}_{ij}}. \label{eq:rr} 
\end{align} \end{subequations} 

\subsection{Evaluation Protocol}\label{sec:method_eval}
For validation, we sampled five (or all available) images per ID and evaluated within-trajectory retrieval performance on the validation set, using cosine similarity as the retrieval score and mAP as the performance metric.
Testing followed the same procedure but across cameras: images from the validation set acted as queries and the test set served as the gallery.

Since no ground-truth cross-camera matches were available, we manually inspected the retrieval results of an early version of the pipeline and manually identified 18 correctly matched fish across the two cameras.
For fairness, we examined results from both the ensemble method (model 10, Table \ref{tab:res}) and the full-image baseline network (model 1, Table \ref{tab:res}). 

To estimate the uncertainty of our calculated test and validation mAP values, we performed a non-parametric bootstrap over the per-query average precision (AP) scores with $B=50,000$ resamples. This yielded 95\% confidence intervals (CI) for all models from the 2.5th and 97.5th percentiles of the bootstrap mAP distributions.
Additionally, we bootstrapped the pairwise differences in mAP between every model pair, and computed two-sided p-values for the null hypothesis $H_0: \Delta mAP = 0$ (APs mean-centered before resampling).
The p-values relevant to our main findings are reported in Section \ref{sec:results} (full set in supplemental materials).
Since the hypothesis tests involve 10 models evaluated on two splits ($2 \cdot \sum_{i=1}^9 i = 90$ comparisons), we controlled the family-wise error rate with a Bonferroni correction, giving a per-test significance level of $\alpha= \frac{0.05}{90} \approx 5.6 \cdot 10^{-4}$.

\section{Results}
\label{sec:results}
Test and validation performance of our models (Table \ref{tab:res}) together with p-values for mAP differences (Section \ref{sec:method_eval}) reveal four main findings.
First, the ensemble approach outperforms the full-image baseline on the test ($p < 10^{-4}$) and validation ($p < 10^{-4}$) sets. The performance gap between the ensemble and baseline is much larger on the test set than on the validation set, indicating that the ensemble approach generalizes substantially better to realistic out-of-distribution conditions. Patch hold-out ablations provided in the supplemental materials confirm that all patches contribute to the final test performance of the ensemble. 
Second, sliced-patch models outperform their corresponding single-patch models on the test (Q1: $p < 10^{-4}$, Q2: $p = 0.199$ not significant [n.s.]) and validation (both $p < 10^{-4}$) sets. 
Third, the full-image ViT model outperforms the ConvNeXt model on the test ($p < 10^{-4}$) and validation ($p < 10^{-4}$) sets. 
Fourth, for both single-patch and sliced-patch networks, the Q2 patch outperforms Q1 by a large margin on the test set (both $p < 10^{-4}$) and by a small margin on the validation set (single: $p=6.8 \cdot 10^{-4}$ n.s., sliced: $p=1.84 \cdot 10^{-2}$ n.s.). 
We hypothesize that this is due to the greater planarity of Q2, which makes its appearance more stable across viewing angles and ensures that training augmentations resemble realistic appearance changes at test time.

\begin{table}[ht]
\centering
\begin{tabular}{|c|c|c|c|}
\hline
\makecell{\#} & Setup & Val mAP $\uparrow$ & Test mAP $\uparrow$\\
\hline
1 & FI (ViT) & 0.932 (0.92, 0.95) & 0.609 (0.54, 0.68) \\
2 & FI (CN) & 0.870 (0.85, 0.89) & 0.270 (0.21, 0.33)\\
\hline
3 & Q1 sgl. & 0.847 (0.82, 0.87) & 0.378 (0.31, 0.45) \\
4 & Q2 sgl. & 0.884 (0.86, 0.90) & 0.718 (0.65, 0.78)\\
5 & Q1 sl. & 0.897 (0.88, 0.92) & 0.573 (0.49, 0.65)\\
6 & Q2 sl. & 0.918 (0.90, 0.94) & 0.754 (0.68, 0.82) \\
7 & head &  0.876 (0.86, 0.89) & 0.517 (0.44, 0.60) \\
8 & dfin. & 0.862 (0.84, 0.88) & 0.531 (0.46, 0.60) \\
\hline
9 & Ens. sgl. & 0.960 (0.95, 0.97) & 0.798 (0.74, 0.85) \\
\rowcolor{gray!20} 10 & Ens. sl. & \textbf{0.965 (0.95, 0.98)} & \textbf{0.860 (0.81, 0.91)} \\
\hline
\end{tabular}
\caption{
mAP results with 95\% CIs for all models. Ensemble sliced (highlighted in gray) achieves the highest performance on both same-trajectory validation and cross-camera test sets. Abbreviations: CN (ConvNeXt), FI (full-image), ens. (ensemble), sl. (sliced), sgl. (single), dfin. (dorsal fin).}\label{tab:res}
\end{table}

\begin{figure}
    \centering
    \includegraphics[width=\linewidth]{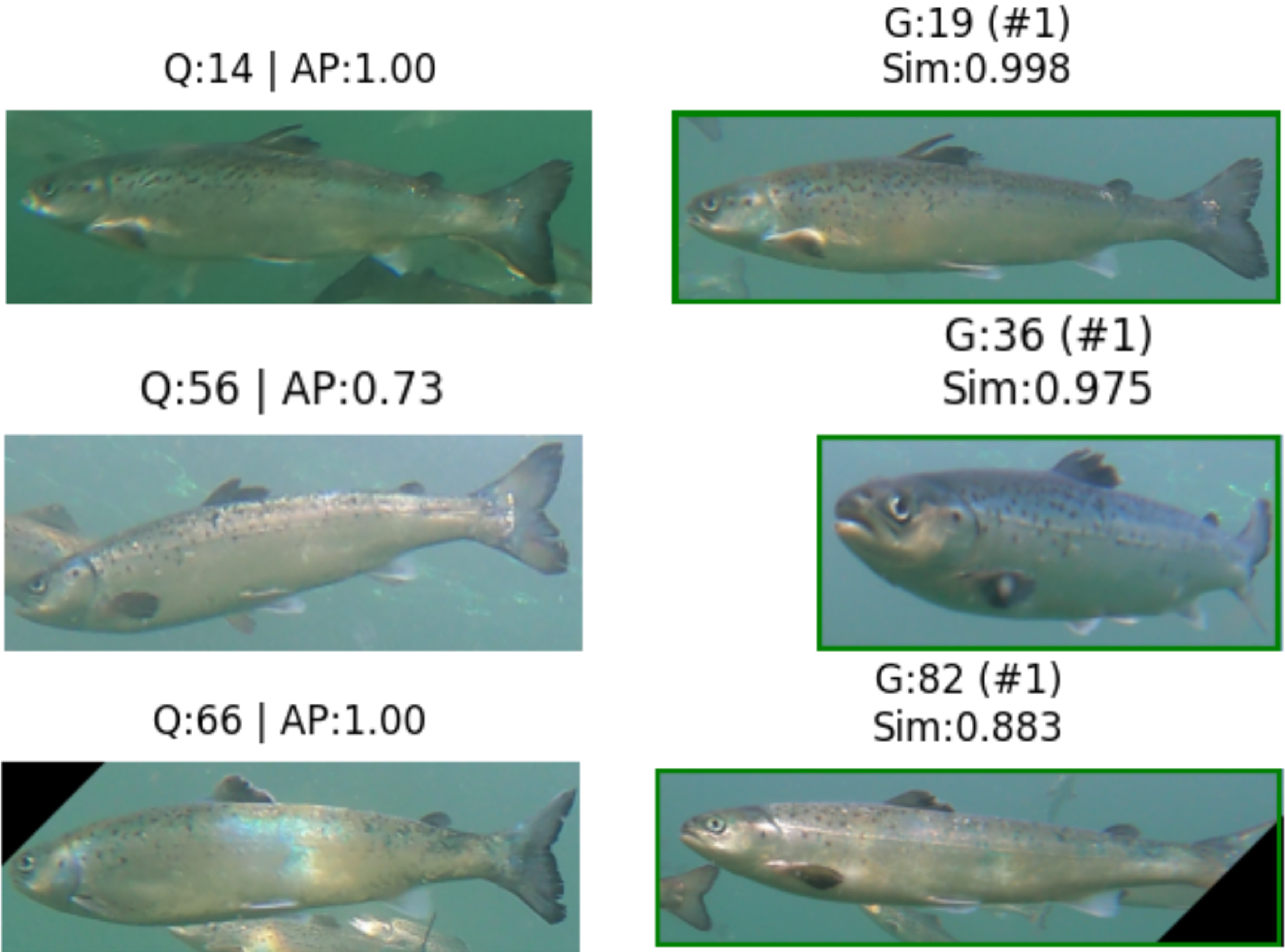}
    \caption{Three challenging cross-camera retrieval examples where the proposed embedding method correctly ranks a correct gallery sample highest. The average precision (AP) is shown for each query. 
    }
    \label{fig:retrieval_exp}
\end{figure}

Figure \ref{fig:retrieval_exp} shows examples of challenging cross-camera retrieval cases where our ensemble approach assigns the highest similarity to a true match, despite variations in lighting (top image), pose (middle image), deformation (bottom image), and geometric distortion (bottom image).

\section{Conclusion}
\label{sec:conclusion}
In this work, we presented a salmon re-identification framework that addresses two central challenges of large-scale salmon re-identification: the lack of labeled data and the high accuracy requirements imposed by commercial net-pens. The lack of labeled data was addressed by using trajectory IDs as proxy labels, which in turn introduced trajectory-ID bias. Both the accuracy requirements and this bias were mitigated through an ensemble approach that divides each salmon into patches and fuses their outputs into a final identity decision. We also introduced an experimental setup using multiple cameras, enabling realistic re-identification evaluation in commercial net-pens. Our results show that the full-image baseline loses 0.323 mAP from validation to cross-camera testing, while our ensemble loses only 0.105 mAP, demonstrating substantially improved robustness.

Our approach has limitations. 
First, we evaluate mAP only on verified true matches, meaning that unmatched fish with high similarity scores do not impact the performance. Future work will explore metrics that quantify the similarity gap between correct and uncertain matches. 
Second, the cross-camera evaluation is based on 18 manually verified matches. While this limits the statistical power of the evaluation, this subset reflects the currently available ground-truth. Results should therefore be interpreted as indicative rather than definitive, and future work will expand the annotated set.
Finally, the multi-network architecture is computationally demanding, and future work will investigate shared backbones and single-pass processing to reduce size and runtime.

Overall, our findings indicate that combining patch ensembles with a multi-camera evaluation setup provides a robust and scalable direction for achieving large-scale salmon re-identification in commercial aquaculture.

\vfill\pagebreak

\bibliographystyle{IEEEbib}
\bibliography{strings,refs}

\newpage

\section{Supplemental materials}
\subsection*{Ablations}
Table \ref{tab:ablation_holdout} presents the
test mAP for model 10 (ensemble sliced) when each patch is removed in turn, showing that the performance decreases when a patch is removed. 

Tables \ref{tab:ablation_lambda}, \ref{tab:ablation_tau} and \ref{tab:ablation_k} report the test mAP for model 10 across different parameter settings. Within the ranges $\lambda \in [0,0.8]$, $\tau \in [0.7,2.0]$, and $k \in [20,500]$ (which are sub-ranges of the parameter sweeps), the mAP varies by less than 0.002. This indicates that the ensemble performance remains stable over a broad range of parameter settings. 

\begin{table}[ht]
\centering
\begin{tabular}{|c|c|}
\hline
\textbf{\makecell{Holdout\\patch}} & \textbf{\makecell{Test\\mAP}} \\
\hline
Q1 & 0.837\\
\hline
Q2 & 0.759\\
\hline
Head & 0.845\\
\hline
Dorsal fin & 0.808\\
\hline
\rowcolor{gray!20} None & 0.860\\
\hline
\end{tabular}
\caption{Holdout ablation. Test mAP for model 10 (ensemble sliced) when each patch is removed in turn.} \label{tab:ablation_holdout}
\end{table}

\begin{table}[ht]
\centering
\begin{tabular}{|c|c|}
\hline
$\boldsymbol{\lambda}$ & \textbf{\makecell{Test\\mAP}} \\
\hline
0 & 0.861\\
\hline
0.2 & 0.861\\
\hline
0.4 & 0.861\\
\hline
0.6 & 0.860\\
\hline
\rowcolor{gray!20} 0.75 & 0.860\\
\hline
0.8 & 0.860\\
\hline
1.0 & 0.844\\
\hline
\end{tabular}
\caption{Lambda ablation. Test mAP for model 10 (ensemble sliced) across different $\lambda$ values. All other parameter values are as described in the main text. The $\lambda$ value used in the paper is highlighted in grey.} \label{tab:ablation_lambda}
\end{table}

\begin{table}[ht]
\centering
\begin{tabular}{|c|c|}
\hline
$\boldsymbol{\tau}$ & \textbf{\makecell{Test\\mAP}} \\
\hline
0.2 & 0.847\\
\hline
0.5 & 0.858\\
\hline
\rowcolor{gray!20} 0.7 & 0.860\\
\hline
1.0 & 0.861\\
\hline
2.0 & 0.861\\
\hline
5.0 & 0.858\\
\hline
\end{tabular}
\caption{Tau ablation. Test mAP for model 10 (ensemble sliced) across different $\tau$ values. All other parameter values are as described in the main text. The $\tau$ value used in the paper is highlighted in grey.} \label{tab:ablation_tau}
\end{table}

\begin{table}[ht]
\centering
\begin{tabular}{|c|c|}
\hline
\textbf{k} & \textbf{\makecell{Test\\mAP}} \\
\hline
1 & 0.817\\
\hline
10 & 0.858\\
\hline
\rowcolor{gray!20} 20 & 0.860\\
\hline
30 & 0.860\\
\hline
60 & 0.861\\
\hline
100 & 0.861\\
\hline
150 & 0.862\\
\hline
200 & 0.861\\
\hline
300 & 0.861\\
\hline
500 & 0.861\\
\hline
\end{tabular}
\caption{k ablation. Test mAP for model 10 (ensemble sliced) across different $k$ values. All other parameter values are as described in the main text. The $k$ value used in the paper is highlighted in grey.} \label{tab:ablation_k}
\end{table}

\FloatBarrier

\subsection*{Statistical Analysis}
\begin{figure}[ht]
    \centering
    \begin{subfigure}[b]{0.5\textwidth}
        \centering
        \includegraphics[width=\textwidth]{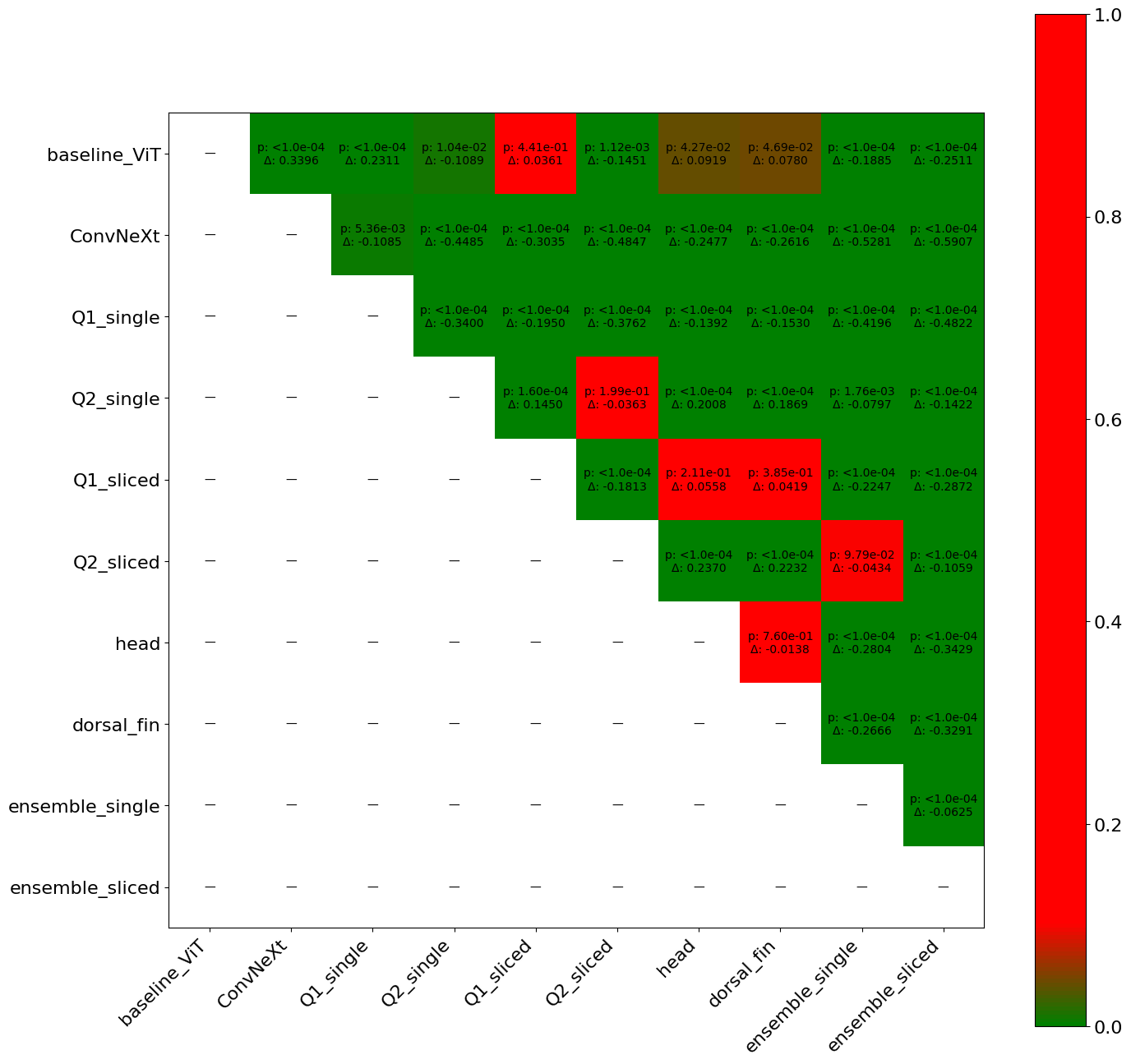}
        \caption{Test}
        \label{fig:pairwise_comp_test}
    \end{subfigure}
    \hfill 
    \begin{subfigure}[b]{0.5\textwidth}
        \centering
        \includegraphics[width=\textwidth]{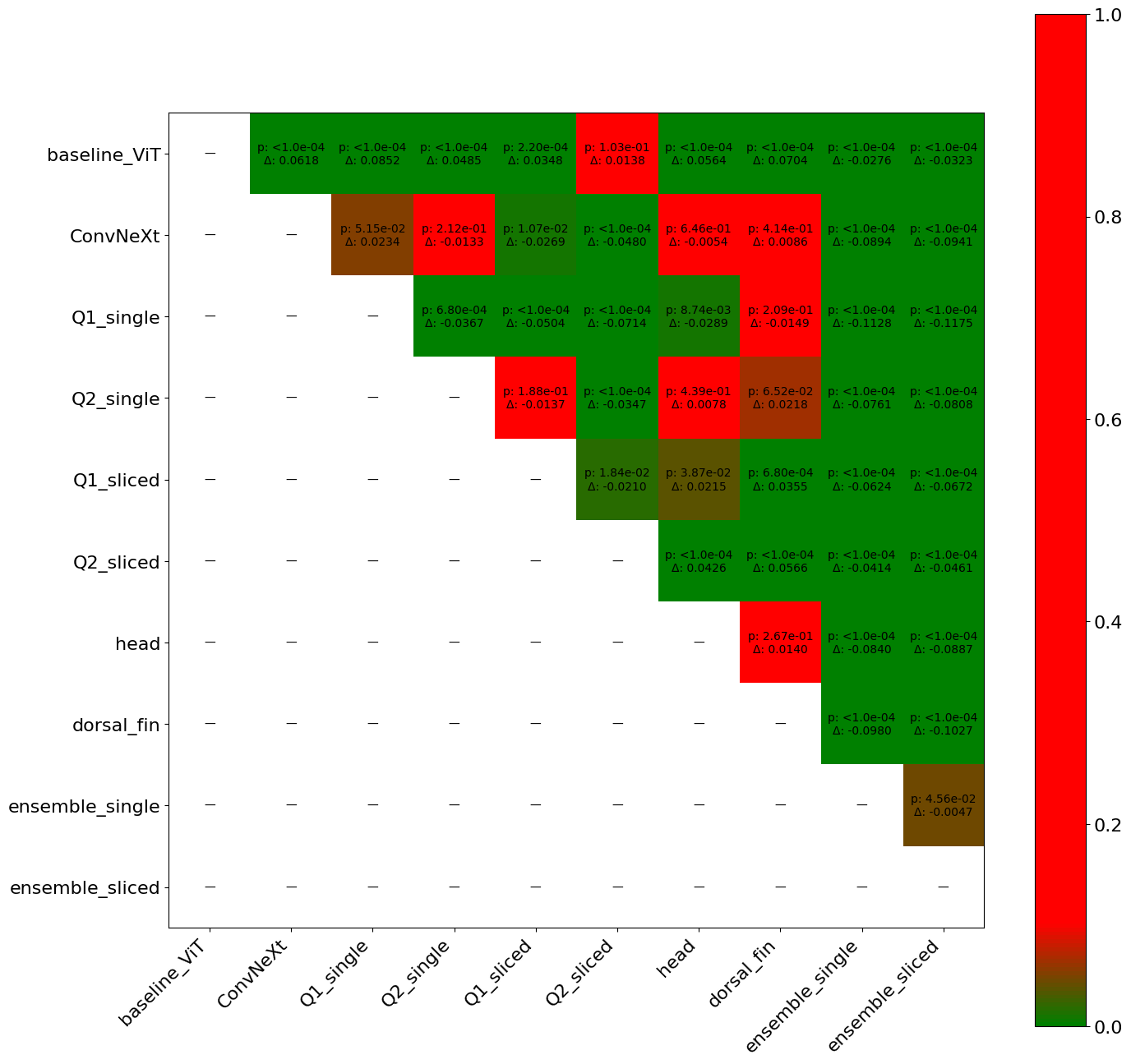}
        \caption{Validation}
        \label{fig:pairwise_comp_val}
    \end{subfigure}
    \caption{Pairwise statistical comparison of test (a) and validation (b) mAP across all 10 models. Each cell shows the two-sided bootstrap p-value and the empirical mAP difference ($\Delta = mAP_{row} - mAP_{column}$) between the row model (left label) and column model (bottom label). Low p-values are marked in green, high p-values are marked in red, and cells that are omitted since they provide no additional information (the diagonal and lower-triangle entries) are displayed in white.}
    \label{fig:pairwise_comp}
\end{figure}

The pairwise differences in mAP (empirical mAP differences and two-sided bootstrap p-values) between all 10 models are shown in Figure \ref{fig:pairwise_comp_test} (test) and Figure \ref{fig:pairwise_comp_val} (validation). The reported p-values are obtained from pairwise model comparisons without accounting for the number of tests performed. To account for the multiple hypotheses, we use a Bonferroni-corrected significance threshold of $\alpha= \frac{0.05}{90} \approx 5.6 \cdot 10^{-4}$.

\end{document}